\makeatletter
\@namedef{ver@fixltx2e.sty}{9999/99/99}
\makeatother
\documentclass[conference,a4paper]{IEEEtran}
\IEEEoverridecommandlockouts

\usepackage[hidelinks]{hyperref}
\usepackage[cmex10]{amsmath}
\usepackage{amssymb,amsfonts}
\usepackage{dblfloatfix}

\usepackage[ruled,vlined]{algorithm2e}

\usepackage{graphicx}
\graphicspath{{Figures/PDF/}{Figures/PNG/}}

\usepackage{booktabs}
\usepackage{tabularx}
\usepackage{array}
\usepackage{siunitx}
\usepackage[numbers,compress]{natbib}
\usepackage{texnames}
\usepackage{bm,bbm}
\usepackage{orcidlink}

\begin{document}

\title{SUMMARIZE FIRST, DOWNLOAD LATER: ONBOARD VLMS FOR BANDWIDTH-EFFICIENT EARTH OBSERVATION}

\author{\IEEEauthorblockN{Junghwan Park \;\; Sangcheol Sim \;\; Woojin Cho \;\; Darongsae Kwon}
\IEEEauthorblockA{
\textit{TelePIX} \\
07330, Seoul, South Korea \\
\{junghwan, sim2real, woojin, darong.kwon\}@telepix.net \\
}
}

\maketitle
\begin{abstract}
Modern Earth observation (EO) satellites carry increasingly advanced sensors that produce vast volumes of high-resolution, multispectral data, yet downlink capacity remains a critical bottleneck — often causing significant latency or the loss of valuable observations within limited contact windows. We propose a ``Summarize First, Download Later'' paradigm that exploits recent advances in onboard edge computing and Vision-Language Models (VLMs). Rather than indiscriminately downlinking raw imagery, the system follows a three-phase interaction protocol: the satellite first transmits concise natural language summaries generated by a quantized onboard VLM; ground operators then issue targeted Visual Question Answering (VQA) queries to verify scene relevance (e.g., wildfires or maritime anomalies); and full-resolution images are downloaded only when critical information is confirmed. This transforms the downlink from passive bulk transfer into an active, semantics-aware dialogue. We implement and evaluate the system on a resource-constrained NVIDIA Jetson platform, and experiments on diverse remote sensing scenes show that the proposed strategy substantially reduces bandwidth consumption while accelerating time-to-insight for time-sensitive missions.
\end{abstract}


\section{Introduction}

Modern Earth observation (EO) satellites continuously acquire high-resolution optical and multispectral imagery, enabling detailed monitoring of the Earth’s surface.
However, satellite-to-ground communication remains constrained by limited downlink bandwidth and short contact windows, particularly for small satellites and commercial constellations.
As sensing capability continues to scale, a growing fraction of collected data cannot be transmitted in a timely manner, even though only a small subset of observations is operationally valuable.

To mitigate this imbalance, recent EO missions increasingly rely on onboard processing to prioritize or filter data before transmission.
This shift toward onboard intelligence, often referred to as orbital edge computing, has been enabled by commercial-off-the-shelf hardware and maturing onboard AI software stacks.
Several missions have demonstrated its effectiveness.
$\Phi$Sat-1~\cite{giuffrida2021varphi} showed that in-orbit neural networks can filter cloudy images, while the WorldFloods demonstration~\cite{mateogarcia2021towards} illustrated that task-specific products such as flood masks can substitute for raw imagery in time-critical scenarios.
ESA’s $\Phi$Sat-2~\cite{guerrisi2023artificial} further extends this direction by supporting multiple onboard AI applications within a unified framework.

\begin{figure}[t]
\centering
\includegraphics[width=1.0\columnwidth]{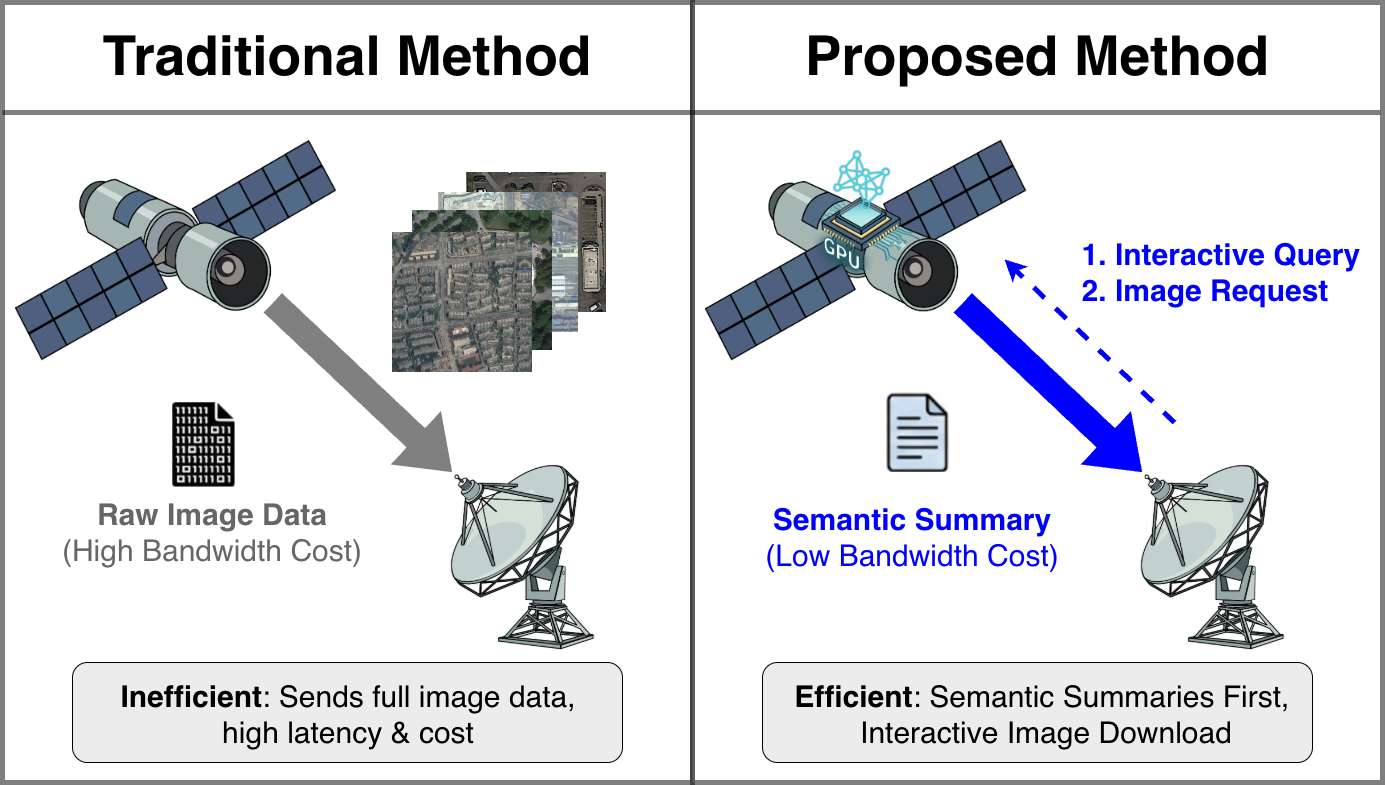}
\caption{Comparison between conventional image-centric downlink and the proposed summarize-first, download-later approach with interactive image requests.}
\label{fig:concept_vis}
\end{figure}

Despite these advances, most existing downlink strategies remain largely image-centric~\cite{furutanpey2025fool}.
In conventional pipelines, satellites transmit full-resolution images whenever contact is available, regardless of scene utility.
As illustrated on the left side of Fig.~\ref{fig:concept_vis}, this approach treats downlink as a one-shot transfer of raw data, leading to high latency and bandwidth cost even when many images are dominated by clouds or benign background.
Although some selective downlink schemes exist~\cite{gomez2024tackling,giuffrida2020cloudscout}, they typically rely on predefined task outputs and provide limited semantic context to ground operators, making it difficult to decide whether an observation truly warrants full image download.

Recent progress in vision-language models (VLMs) offers a new opportunity to rethink this paradigm~\cite{radford2021learning,li2023blip}.
By mapping high-dimensional visual inputs into natural language, VLMs enable compact semantic representations that are directly interpretable by human operators.
Unlike fixed task-specific scores or masks, natural-language summaries can flexibly convey scene context, salient events, and uncertainty, while also supporting interactive querying.

Building on this capability, we propose a Summarize First, Download Later paradigm for bandwidth-efficient EO.
As illustrated on the right side of Fig.~\ref{fig:concept_vis}, the satellite first generates concise semantic summaries of captured multispectral imagery using an onboard VLM and transmits them at minimal bandwidth cost.
Based on these summaries, the ground station can interactively issue targeted queries to validate or refine the onboard interpretation.
Full-resolution imagery, or selected regions of interest, is downloaded only after semantic relevance has been confirmed.
This interaction-driven workflow enables flexible and informed downlink decisions, transforming satellite communication from bulk data transfer into a semantics-aware process.

This paper makes the following contributions:
\begin{itemize}

\item \textbf{Interaction-driven downlink paradigm.}
We introduce a summarize-first, download-later transmission framework that reformulates satellite downlink as an interactive, semantics-driven decision process rather than a one-shot image transfer.

\item \textbf{Onboard vision-language interaction.}
We leverage onboard vision-language models to generate lightweight semantic summaries and support interactive question answering, enabling human-in-the-loop assessment of scene relevance before committing downlink bandwidth.

\item \textbf{Onboard prototype and feasibility analysis.}
We implement an end-to-end prototype on resource-constrained onboard hardware and provide a quantitative feasibility analysis that characterizes (i) semantic fidelity of text-first summaries, (ii) runtime and memory cost of onboard VLM inference, and (iii) potential bandwidth reduction achievable by deferring image downlink in favor of text-first transmission.

\end{itemize}

\vspace{0.3em}

\section{Background and Related Work}\label{sec:related}

\subsection{Onboard Processing for Earth Observation Satellites}
Onboard processing for Earth observation (EO) satellites is driven by the growing mismatch between sensing capability and limited downlink capacity~\cite{furutanpey2025fool}.
High-resolution optical and multispectral sensors generate data volumes that cannot be fully transmitted under short contact windows, motivating onboard intelligence that can prioritize high-value observations.

Recent missions have demonstrated the feasibility of this paradigm.
$\Phi$Sat-1~\cite{giuffrida2021varphi} validated in-orbit neural inference for cloud filtering, while $\Phi$Sat-2~\cite{guerrisi2023artificial} extends this concept toward a more general onboard AI application stack.
Rather than fully autonomous operation, several studies emphasize \emph{human-in-the-loop} approaches~\cite{tuia2023artificial}, where onboard analytics provide preliminary interpretations that are validated or refined on the ground.

Existing selective downlink strategies~\cite{ruuvzivcka2022ravaen, chatar2023data} typically rely on task-specific scores or predefined outputs such as masks or alerts.
While effective, these approaches are limited in flexibility.
This motivates interaction-aware transmission schemes that can convey semantic information in a compact and human-interpretable form, enabling informed downlink decisions without immediate image transfer.

\subsection{Foundation Models for Remote Sensing}
Vision-language models (VLMs) enable semantic compression by distilling high-dimensional visual inputs into natural language representations.
Recent progress in general-purpose VLMs includes instruction-tuned multimodal models~\cite{liu2023visual}, image-language pretraining frameworks such as BLIP-2~\cite{li2023blip2}, and open-source backbones including Qwen-VL~\cite{bai2023qwenvl}.

In remote sensing, captioning datasets such as RSICD~\cite{lu2017exploring} and satellite-based visual question answering benchmarks~\cite{lobry2020rsvqa,yuan2022easy} demonstrate that EO imagery can be effectively mapped to linguistic descriptions and queried interactively.
However, most prior work~\cite{zi2025rsvlm, li2024vrsbench} treats language outputs as an offline analysis tool rather than as part of an operational satellite-to-ground communication loop.

In contrast, this paper leverages VLM-generated text as a lightweight semantic interface for onboard decision making and downlink control.
By using natural language summaries and question-answering as first-class communication primitives, the proposed approach bridges foundation models and practical, bandwidth-efficient EO operations.

\section{System and Protocol}\label{sec:system}

\begin{figure}[t]
\centering
\includegraphics[width=1.0\columnwidth]{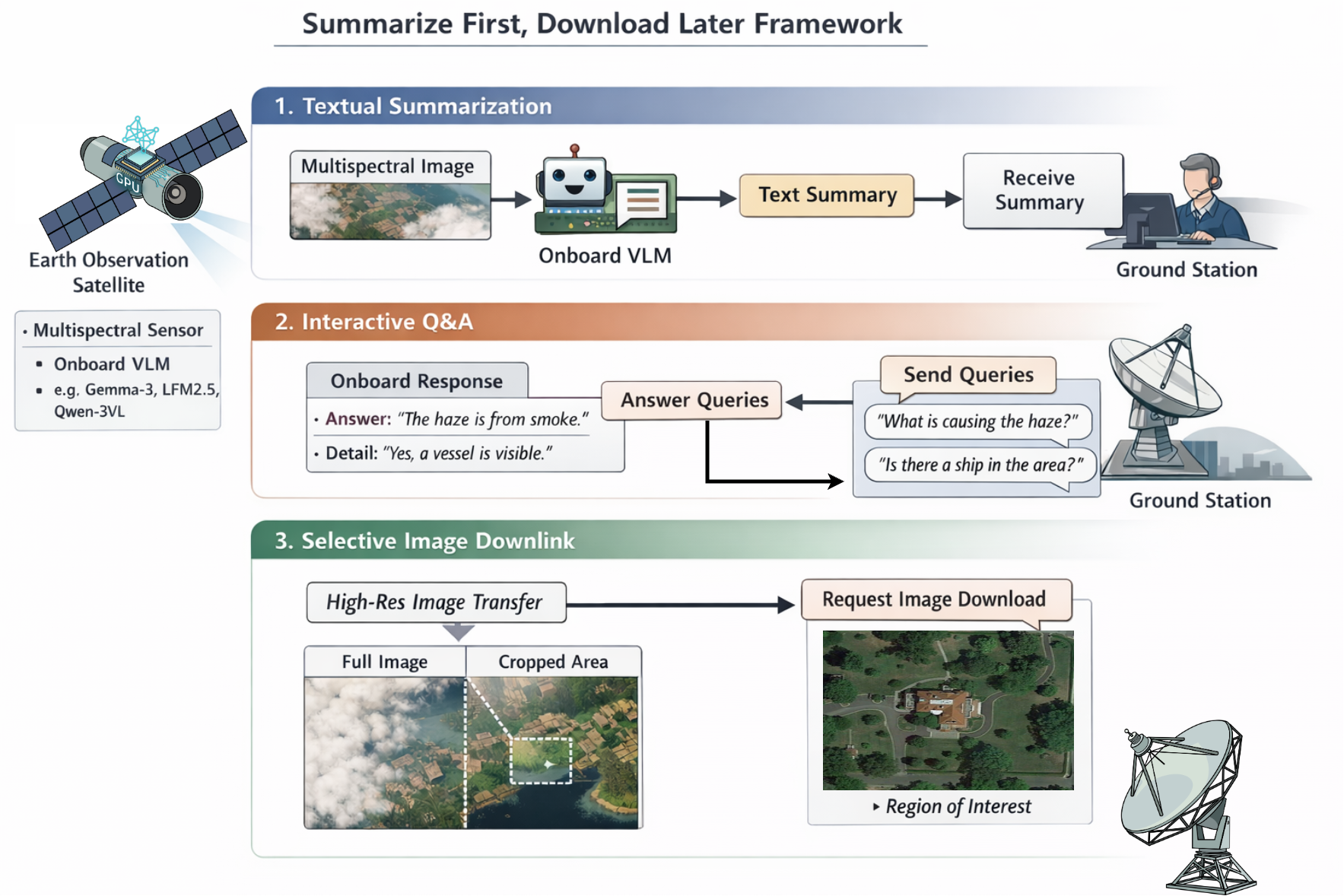}
\caption{Three-phase onboard--ground interaction pipeline used in experiments.}
\label{fig:pipeline}
\end{figure}

We propose a three-phase onboard--ground interaction protocol designed to refine information progressively before committing high-bandwidth image transmission. As illustrated in Fig.~\ref{fig:pipeline}, the key idea is to treat downlink not as a one-shot data dump, but as an interactive decision process guided by semantic understanding. This unified pipeline governs all vision-language tasks in our experiments, ensuring that both captioning and VQA contribute jointly to informed downlink decisions.

The proposed framework operates through the following three distinct phases:

\begin{itemize}
    \item \textbf{Phase 1: Textual Summarization.}
    For each captured multispectral image, the satellite generates a lightweight semantic packet using an onboard vision-language model. Instead of transmitting raw pixel data, this packet includes a concise natural-language description of the scene. These text-first packets are transmitted frequently using minimal bandwidth, enabling the ground system to rapidly survey onboard observations and obtain a compact semantic overview without the latency of full image downlink.

    \item \textbf{Phase 2: Interactive Question Answering.}
    Based on the received summaries, ground operators may issue targeted natural-language queries back to the satellite. This phase serves as a validation step to resolve uncertainty or refine the initial interpretation. Typical queries include checking for specific phenomena (e.g., smoke, ships, flooding) or requesting localized descriptions of regions of interest. The satellite responds with short textual answers, allowing operators to confirm data utility through a low-bandwidth VQA interface.

    \item \textbf{Phase 3: Selective Image Downlink.}
    Full-resolution imagery or specific crops are requested only after semantic relevance has been confirmed via the previous phases. By deferring bulk data transfer until after interactive validation, the protocol ensures that scarce downlink bandwidth is allocated exclusively to high-value observations.
\end{itemize}

This progressive, interaction-driven design transforms satellites from passive sensors into intelligent agents. By combining onboard semantic understanding with ground-in-the-loop validation, the system effectively minimizes unnecessary transmissions while preserving access to mission-critical imagery when it matters most.

\section{Experimental Setup}\label{sec:experiment}

We evaluate the proposed summarize-first, download-later paradigm using real remote sensing imagery and onboard inference on resource-constrained hardware.
All experiments are designed to reflect realistic satellite operation scenarios, where both computation and communication budgets are limited.

\subsection{Onboard Hardware Configuration}
All vision-language inference is performed on an NVIDIA Jetson-class embedded platform, representative of contemporary onboard AI payloads.
The platform executes image preprocessing, multimodal inference, and text generation fully onboard, without offloading intermediate representations.
This setup allows us to assess both feasibility and latency under practical onboard constraints.

\subsection{Vision-Language Models}
We benchmark three compact vision-language models suitable for embedded, resource-constrained deployment.
The selected models differ in parameter scale and multimodal design, enabling a diverse comparison of onboard VLM capabilities.

\begin{itemize}
    \item \textbf{Gemma-3 4B (4-bit)}~\cite{gemma_2025,team2025gemma}  
    A quantized multimodal model from the Gemma family developed by Google DeepMind, supporting joint image–text understanding with an emphasis on efficient on-device inference.

    \item \textbf{LFM 2.5 VL 1.6B (8-bit)}~\cite{liquidai2025lfm2}  
    A lightweight vision-language model from the LFM2.5 series, designed for embedded multimodal reasoning with a compact footprint and instruction tuning.

    \item \textbf{Qwen3-VL 2B (8-bit)}~\cite{qwen3technicalreport}  
    A recent vision-language model from the Qwen family, providing strong visual question answering and captioning performance at a moderate parameter scale.
\end{itemize}

All models are evaluated in inference-only mode on remote sensing imagery without task-specific finetuning.

\subsection{Datasets and Tasks}
Experiments focus on two core semantic understanding tasks that are directly relevant to interactive downlink decisions: visual question answering and image captioning. All datasets consist of real remote sensing imagery commonly used in vision--language research for Earth observation.

\begin{itemize}
    \item \textbf{Visual Question Answering (VQA).}
    We evaluate the ability of onboard vision--language models to answer targeted natural-language queries about satellite images.
    Experiments are conducted on two remote sensing VQA datasets, referred to as RSVQA-LR and RSVQA-HR~\cite{lobry2020rsvqa}.
    RSVQA-LR is constructed from Sentinel-2 satellite imagery at relatively low spatial resolution, while RSVQA-HR is based on higher-resolution aerial imagery.
    Both datasets cover a wide range of geographic scenes, including urban areas, agricultural land, coastal regions, forests, and water bodies.
    The questions include object presence, land-use classification, counting, and scene-level semantic attributes.

    \item \textbf{Image Captioning.}
    We assess whether compact textual summaries can convey scene semantics prior to image downlink.
    Two remote sensing image captioning datasets are used: RSICD~\cite{lu2017exploring} and NWPU-Captions~\cite{cheng2022nwpu_captions}.
    RSICD consists of high-resolution satellite images annotated with multiple human-written captions describing land-cover composition and spatial context.
    NWPU-Captions contains a larger collection of aerial images with diverse scene types, including residential areas, transportation infrastructure, agricultural regions, and natural landscapes.
    Captions describe dominant objects, land-cover categories, and spatial relationships within each scene.
\end{itemize}

Across both tasks, datasets are intentionally kept small and representative, reflecting the practical constraints of onboard validation rather than large-scale offline benchmarking.



\section{Quantitative Results}\label{sec:quantitative}

We first present quantitative results for onboard VQA and image captioning, focusing on accuracy-oriented metrics and relative performance trends across models.

\subsection{Visual Question Answering Performance}
Table~\ref{tab:vqa_results} reports VQA accuracy for each onboard vision-language model across two remote sensing datasets.
Despite operating under strict resource constraints, all models demonstrate meaningful semantic reasoning capability.
Performance varies depending on scene complexity and question type, with higher accuracy observed for object presence and coarse land-use queries.

\begin{table}[t]
\centering
\caption{Onboard VQA performance.}
\label{tab:vqa_results}
\begin{tabular}{lcc}
\toprule
Model & RSVQA-LR & RSVQA-HR \\
\midrule
Gemma3 & 72.2 & 61.0 \\
LFM2.5 & 69.3 & 52.8 \\
Qwen3VL & 73.6 & 62.5 \\
\bottomrule
\end{tabular}
\end{table}

These results suggest that lightweight VLMs can support interactive semantic validation prior to image downlink, particularly for high-level situational awareness.

\subsection{Image Captioning Metrics}
Table~\ref{tab:caption_results} summarizes captioning performance using standard language similarity metrics.
Specifically, Gemma3 achieves the highest BERTScore-F1 on the NWPU dataset (0.901), indicating strong lexical match with reference texts, while Qwen3VL consistently outperforms others in CLIPScore, particularly on RSICD (0.314), suggesting superior visual-semantic alignment.
Despite these minor variations, all evaluated models maintain competitive performance within a narrow margin.
While absolute scores remain lower than those reported in large-scale offline benchmarks, the generated captions are sufficiently informative for human-in-the-loop decision making.

\begin{table}[t]
\centering
\caption{Onboard image captioning results. We report BERTScore-F1 and CLIPScore (higher is better).}
\label{tab:caption_results}
\resizebox{\columnwidth}{!}{%
\begin{tabular}{lcccc}
\toprule
 & \multicolumn{2}{c}{NWPU} & \multicolumn{2}{c}{RSICD} \\
\cmidrule(lr){2-3}\cmidrule(lr){4-5}
Model & BERTScore-F1 & CLIPScore & BERTScore-F1 & CLIPScore \\
\midrule
Gemma3 & \textbf{0.901} & 0.286 & \textbf{0.894} & 0.298 \\
LFM2.5 & 0.899 & 0.282 & 0.893 & 0.300 \\
Qwen3VL & 0.895 & \textbf{0.295} & 0.886 & \textbf{0.314} \\
\bottomrule
\end{tabular}}
\end{table}

Importantly, these captions can be transmitted at negligible bandwidth cost compared to image payloads, aligning with the goals of the proposed protocol.

\subsection{Bandwidth Savings}\label{sec:bandwidth}
We quantify bandwidth efficiency by comparing the size of the proposed semantic text packet against standard image payloads.
As shown in Table~\ref{tab:bw_savings}, the Phase-1 packet is a compact UTF-8 JSON message with a mean size of approximately \SI{485}{B}.
In comparison, the compressed PNG \emph{preview} images used in our experiments average roughly \SI{100}{kB}, representing a data overhead of more than \(200\times\) relative to the text packet.
When considering a standard full-resolution EO image (\(\approx\)\SI{50}{MB}), this reduction factor increases to approximately \(1.0\times10^5\).
These ratios demonstrate that the proposed text-first approach significantly reduces immediate downlink requirements, allowing for frequent semantic communication even under severe bandwidth limitations.

\begin{table}[t]
\centering
\caption{Bandwidth savings from text-first transmission.}
\label{tab:bw_savings}
\begin{tabular}{lrr}
\toprule
Payload type & Size & Reduction vs.\ packet \\
\midrule
Phase-1 text packet & \SI{485}{B} & \(1\times\) \\
RSICD preview image & \SI{97}{kB} & \(\approx 2.0\times10^2\) \\
NWPU preview image & \SI{105}{kB} & \(\approx 2.2\times10^2\) \\
Illustrative full-res EO image & \SI{50}{MB} & \(\approx 1.0\times10^5\) \\
\bottomrule
\end{tabular}
\end{table}

\subsection{Onboard Runtime Benchmark (Jetson)}
To quantify feasibility on embedded hardware, we benchmark end-to-end multimodal inference on a Jetson Orin Nano. Table~\ref{tab:jetson_bench} reveals clear trade-offs between runtime and memory footprint. LFM2.5 achieves the fastest inference (29.06\,s) and smallest peak memory (1.6\,GB), driven largely by its short loading time (3.35\,s) — well suited to frequent, low-latency onboard semantic checks. Gemma3, in contrast, incurs the longest wall-clock time (106.79\,s) and highest memory use (3.4\,GB), indicating that initialization dominates its end-to-end latency. Qwen3VL strikes a balance, with moderate runtime and memory while maintaining stable token throughput. Overall, compact VLMs are feasible on Jetson-class platforms, but fast loading and low memory footprint are critical for responsive text-first summarization under constrained onboard conditions.

\begin{table}[t]
\centering
\caption{Jetson runtime benchmark.}
\label{tab:jetson_bench}
\resizebox{\columnwidth}{!}{%
\begin{tabular}{lcccc}
\toprule
Model & Wall(s) & Peak RSS(MB) & Load(s) & Tok/s \\
\midrule
Gemma3 & 106.79 & 3464 & 43.18 & 15.51 \\
LFM2.5 & 29.06 & 1619 & 3.35 & 19.93 \\
Qwen3VL & 51.62 & 2200 & 14.28 & 16.51 \\
\bottomrule
\end{tabular}}
\end{table}

\section{Discussion and Conclusion}\label{sec:conclusion}
This paper presented a summarize-first, download-later paradigm that reframes satellite downlink as a semantics-driven, interactive process. By running vision-language models onboard and transmitting text before imagery, the system enables informed human-in-the-loop decisions under severe bandwidth constraints. Our experiments show that compact VLMs on embedded hardware can generate meaningful captions and answer targeted questions on remote sensing imagery. While not a replacement for full-resolution analysis, they serve as an effective semantic gatekeeper that reduces unnecessary transmission.
Open challenges include onboard robustness to domain shift, uncertainty-aware generation, and energy-efficient multimodal inference. Still, the framework highlights the potential of language as a first-class communication modality for EO satellites, and we expect interaction-driven downlink protocols to grow in importance as constellation scale and sensing resolution increase.

\clearpage

\small
\bibliographystyle{IEEEtranN}
\bibliography{references}

\end{document}